\documentclass[conference]{IEEEtran}
\IEEEoverridecommandlockouts

\usepackage{amsmath}
\usepackage{amsthm}
\usepackage{amsfonts}
\usepackage{amssymb}
\usepackage{graphicx}
\usepackage{mathrsfs}
\usepackage{booktabs}
\usepackage{float}
\usepackage{tikz}
\usepackage{tikz-network}
\usepackage{pgfplots}
\usepackage{enumitem}
\usepackage{caption}
\usepackage{titlesec}
\setlist[description]{leftmargin=*}
\titlespacing*{\section}{0pt}{0.3\baselineskip}{0.1\baselineskip}
\titlespacing*{\subsection}{0pt}{0.3\baselineskip}{0.1\baselineskip}

\newcommand{\tsf}[1]{{\text{\textsf{#1}}}}

\theoremstyle{definition}
\newtheorem{definition}{Definition}
\usetikzlibrary{decorations.pathreplacing,calc,shapes,positioning,arrows,arrows}
\usepackage{pgfplots}
\usepackage{subfigure}
\pgfplotsset{%
	,compat=1.12
	,every axis x label/.style={at={(current axis.right of origin)},anchor=north west}
	,every axis y label/.style={at={(current axis.above origin)},anchor=north east}
}
\usetikzlibrary{datavisualization,shapes.geometric,arrows.meta,decorations.markings, fit}
\usetikzlibrary{datavisualization.formats.functions}
\definecolor{scarlet}{rgb}{1.0, 0.13, 0.0}
\definecolor{brightmaroon}{rgb}{0.76, 0.13, 0.28}
\definecolor{mediumturquoise}{rgb}{0.28, 0.82, 0.8}
\definecolor{fandango}{rgb}{0.71, 0.2, 0.54}
\definecolor{antiquewhite}{rgb}{0.98, 0.92, 0.84}
\definecolor{babyblue}{rgb}{0.54, 0.81, 0.94}
\definecolor{brilliantlavender}{rgb}{0.96, 0.73, 1.0}
\definecolor{bronze}{rgb}{0.8, 0.5, 0.2}
\definecolor{cornsilk}{rgb}{1.0, 0.97, 0.86}
\definecolor{lavenderpink}{rgb}{0.98, 0.68, 0.82}
\definecolor{sandybrown}{rgb}{0.96, 0.64, 0.38}
\definecolor{celadon}{rgb}{0.67, 0.88, 0.69}
\newcommand{\blue}[1]{{\color{blue} #1}}

\newcommand{\grande}{GRANDE }

\newcommand{\best}[1]{$\mathbf{#1}$}
\newcommand{\recall}[1]{$\tsf{r@p}#1$}

\begin{document}
    \title{\grande: a neural model over directed multigraphs with application to anti-money laundering}
    
    \makeatletter
    \newcommand{\linebreakand}{%
      \end{@IEEEauthorhalign}
      \hfill\mbox{}\par
      \mbox{}\hfill\begin{@IEEEauthorhalign}
    }
    \makeatother
    
    \author{\IEEEauthorblockN{Ruofan Wu$^*$}
    \IEEEauthorblockA{
    Ant Group\\
    ruofan.wrf@antgroup.com}
    \and
    \IEEEauthorblockN{Boqun Ma$^*$\thanks{$^*$ Equal contribution}}
    \IEEEauthorblockA{
    Ant Group\\
    boqun.mbq@antgroup.com}
    \linebreakand
    \IEEEauthorblockN{Hong Jin}
    \IEEEauthorblockA{
    Ant Group\\
    jinhong.jh@antgroup.com}
    \and
    \IEEEauthorblockN{Wenlong Zhao}
    \IEEEauthorblockA{
    Ant Group\\
    chicheng.zwl@antgroup.com}\\
    \IEEEauthorblockN{Tianyi Zhang}
    \IEEEauthorblockA{
    Ant Group\\
    zty113091@antgroup.com}
    \and
    \IEEEauthorblockN{Weiqiang Wang}
    \IEEEauthorblockA{
    Ant Group\\
    weiqiang.wwq@antgroup.com}
    }
    \maketitle
    
    \begin{abstract}
        The application of graph representation learning techniques to the area of financial risk management (FRM) has attracted significant attention recently. 
        However, directly modeling transaction networks using graph neural models remains challenging: Firstly, transaction networks are \emph{directed multigraphs} by nature, which could not be properly handled with most of the current off-the-shelf graph neural networks (GNN). Secondly, a crucial problem in FRM scenarios like anti-money laundering (AML) is to identify risky transactions and is most naturally cast into an \emph{edge classification} problem with \emph{rich} edge-level features, which are not fully exploited by the prevailing GNN design that follows node-centric message passing protocols. 
        In this paper, we present a systematic investigation of design aspects of neural models over directed multigraphs and develop a novel GNN protocol that overcomes the above challenges via efficiently incorporating directional information, as well as proposing an enhancement that targets edge-related tasks using a novel message passing scheme over an extension of edge-to-node dual graph. 
        A concrete GNN architecture called \grande is derived using the proposed protocol, with several further improvements and generalizations to temporal dynamic graphs. We apply the \grande model to both a real-world anti-money laundering task and public datasets. Experimental evaluations show the superiority of the proposed \grande architecture over recent state-of-the-art models on dynamic graph modeling and directed graph modeling. 
    \end{abstract}
    % \keywords{graph representation learning, financial risk management, edge representation, edge-to-node duality}
    
    \section{Introduction}
    Recent years have witnessed an increasing trend of adopting modern machine learning paradigms to financial risk management (FRM) scenarios \cite{akib2020ML4fin}. As a typical use case in operational risk scenarios like fraud detection and anti-money laundering, the identification of risky entities (user accounts or transactions) is cast into a supervised classification problem using behavioral data collected from the operating financial platform \cite{chen2018machine, kute2021aml}. For institutions like commercial banks and online payment platforms, the most important source of behavior information is the \emph{transaction records} between users. 
    % Intuitively, the existence of transactions between users often implies a correlation between their behaviors. Moreover, for transactions triggered by the same user, it is also reasonable to consider them as being correlated. The complex correlations in such kinds of financial data make graphical modeling \cite{wang2021review} a powerful alternative to i.i.d. approaches in machine learning. \par 
    % \modification{
    making \emph{transaction networks} (with users as nodes and transactions as edges) a direct and appropriate data model. Unlike standard pattern recognition tasks like image recognition where decisions are made according to information of individual objects, identification of risky patterns over transaction network requires reasoning beyond any individual scope. The phenomenon is particularly evident in the area of anti-money laundering (AML), where suspicious transactions are usually related by several users or accounts, with transactions between them being highly correlated, thereby exhibiting a cascading pattern which makes i.i.d. approaches in machine learning unsuitable.
    The surging developments of machine learning models over graphs, especially graph representation learning \cite{hamilton2020graph}, have attracted significant attention in the financial industry and have shown promising results in the area of FRM \cite{liu2018heterogeneous, liu2021intention}. The dominant practice in graph representation learning is to utilize the panoply of graph neural networks (GNN) \cite{battaglia2018relational} that produce node-level representations via principled aggregation mechanisms which are generally described via message passing protocols \cite{pmlr-v70-gilmer17a} or spectral mechanisms \cite{kipf2016semi}. \par
    Despite their convincing performance, the majority of the existent GNN models operate over \emph{undirected graphs}, which makes them inadequate for the direct modeling of transaction networks. Firstly, many graphs that arise in FRM applications are directed by nature: i.e., in the case of a transaction network with users as nodes and transactions as edges, the direction of an edge is typically understood as the direction of its corresponding cash flow. In areas like anti-money laundering (AML), directional information is generally perceived to be of significant importance and shall not be neglected \cite{Sudjianto2010Statistical}. Secondly, there might exist multiple transactions between certain pairs of users. Thirdly, transactions are naturally associated with timestamps that indicate the time of occurrence. Therefore, to fully utilize the graphical structure of transaction networks, we need representation learning frameworks that support \emph{temporal directed multigraphs}. While recent progress on \emph{dynamic graph neural networks} \cite{xu2020inductive, kazemi2020representation} provide appropriate methods to handle temporality, discussions over neural architectures that supports directed multigraphs remains nascent \cite{pmlr-v70-gilmer17a, ma2019spectral, tong2020directed, NEURIPS2020_cffb6e22, zhang2021magnet}.\par
    From a practical point of view, the targeted risky entities may be either node (i.e., malicious users) or edges (i.e., suspicious transactions). 
    % Conventional GNN architectures excel in learning node representations, and edge representations are usually derived via combining node representations corresponding to both ends of edges. 
    % \modification{
    Conventional GNN architectures produce node-level representations via encoding information of each node's rooted subtrees \cite{xu2018how}, making them a good fit for \emph{user or account level} risk identifications. When the underlying task is to detect risky transactions, the prevailing practice is to present edges using a combination of node representations corresponding to both ends of edges. While such design may be adequate for tasks like link prediction, it lacks a way to effectively integrate edge-level information into the edge representation. Since financial networks usually contain rich edge-level features (i.e., detailed transaction-related information), refinements on edge-level representations are needed. For example, to accurately represent a transaction, we need to combine the information of its buyer (cash sender) and seller (cash receiver), and the transaction-related information, with each of them requiring aggregating relevant information from related users and transactions.
    \begin{figure}
        \centering
        \scalebox{.6}{\begin{tikzpicture}
    \Vertex[label = $ n_0 $, color=celadon, x=0, y=0]{n0},
    \Vertex[label = $ n_1 $, color=antiquewhite, x=-2, y=0]{n1},
    \Vertex[label = $ n_2 $, color=antiquewhite, x=-1.44, y=1.44]{n2},
    \Vertex[label = $ n_3 $, color=antiquewhite, x=0, y=2]{n3},
    \Vertex[label = $ n_4 $, color=antiquewhite, x=1.44, y=1.44]{n4},
    \Vertex[label = $ n_5 $, color=antiquewhite, x=2, y=0]{n5},
    \Vertex[label = $ n_6 $, color=antiquewhite, x=1.44, y=-1.44]{n6},
    \Vertex[label = $ n_7 $, color=antiquewhite, x=0, y=-2]{n7},
    \Vertex[label = $ n_8 $, color=antiquewhite, x=-1.44, y=-1.44]{n8},
    \Vertex[label = $ n_9 $, x=3.44, y=1.44]{n9},
    \Vertex[label = $ n_{10} $, x=3.44, y=-1.44]{n10},
    \Edge[Direct, label=$ e_0 $](n1)(n0)
    \Edge[Direct, label=$ e_1 $](n2)(n0)
    \Edge[Direct, label=$ e_2 $](n3)(n0)
    \Edge[Direct, label=$ e_3 $](n4)(n0)
    \Edge[Direct, label=$ e_4 $](n5)(n0)
    \Edge[Direct, label=$ e_5 $](n6)(n0)
    \Edge[Direct, label=$ e_6 $](n7)(n0)
    \Edge[Direct, label=$ e_7 $](n8)(n0)
    \Edge[Direct, label=$ e_8 $](n5)(n9)
    \Edge[Direct, label=$ e_9 $](n5)(n10)
\end{tikzpicture}}
        \caption{Illustration on the deficiency of the directed message passing protocol in \cite{battaglia2018relational}: suppose the node of interest is $ n_0 $, using GNNs designed according to the protocol in \cite{battaglia2018relational}, it becomes impossible for $ n_0 $ to aggregate information of $ n_9 $ and $ n_{10} $. Under the context of financial risk management, suppose $ n_9 $ and $ n_{10} $ corresponds to known fraudsters, and edges correspond to transactions. Although the riskiness of $ n_5 $ might be undetermined, the transaction pattern makes it highly suspicious and therefore uplifts the riskiness of $ n_0 $. To build models that behave coherently with the above reasoning process, GNN protocols that aggregates information from \emph{both directions} are required }
        \label{fig:digraph_illustration}
    \end{figure}
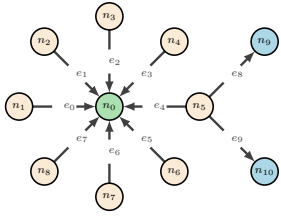
    A recent line of work \cite{jiang2020censnet, cai2020line, jo2021edge} focused directly on \emph{learned edge representations} using the idea of \emph{edge-to-node duality} and obtained satisfactory performance over downstream tasks like edge classification. However, previous works on edge representation learning all applies to undirected graphs, making the extension to transaction networks highly non-trivial.\par
    In this paper, we propose a general message passing neural network protocol that simultaneously outputs node and edge representations over directed multigraphs. Based on this protocol, we derive a GNN architecture called GRANDE with an extension to temporal graphs that efficiently leverages the underlying structural property of transaction networks. More specifically, we summarize our contribution as follows:
    \begin{itemize}[leftmargin=*]
        \item We develop a novel bi-directional message passing protocol with duality enhancement (BiMPNN-DE) that strengthens previous proposals over message passing neural architectures over directed multigraphs. The improvement is two-fold: Firstly, it effectively combines neighborhood information from both incoming and outgoing edges of nodes. Secondly, it simultaneously outputs node and edge representations via performing message passing over both the original graph and its \emph{augmented} edge adjacency graph. 
        \item We derive a concrete GNN architecture following the proposed BiMPNN-DE protocol called GRANDE, that devices the acclaimed transformer \cite{vaswani2017attention} mechanism for neighborhood aggregation. The proposed GRANDE framework is made compatible with temporal directed multigraphs through the integration of a generic time encoding module that further extends previous works on dynamic graph modeling \cite{xu2020inductive}.
        \item To show the practical effectiveness of GRANDE, we apply it to a suspicious transaction identification task in anti-money laundering, with the underlying transaction network data collected from one of the world's leading online payment platforms. Comparisons against various undirected and directed GNN baselines show the superiority of the proposed model. We also provide evaluations on two public datasets generated from transaction networks to further verify the strength of GRANDE framework when underlying graph features are relatively weak.
    \end{itemize}
    
    \section{Methodology}
    \subsection{Problem formulation}\label{sec:problem_formulation}
    Under the context of financial risk management, we consider the following \emph{event stream} representation of recorded transaction data that are available in most online transaction systems: 
    \begin{align}
        \mathcal{E} = \{(u_1, v_1, t_1, \chi_1), (u_2, v_2, t_2, \chi_2), \ldots \}
    \end{align}
    Each event $ (u, v, t, \chi) $ is interpreted as a transaction from user $u$ to user $v$ that occurred at time $t$, with related features $\chi$ that could often be further decomposed as user-level features like user account information, and event-level features like transaction amount and channels. In this paper, we focus on the representative task of \emph{transaction property prediction} that typically takes the form of binary classification that aims at identifying illicit or fraudulent transactions. The task could be cast into a graph learning problem of edge classification in a straightforward manner. We consider the temporal graph modeling paradigm \cite{kazemi2020representation} that views the underlying temporal graph as being generated from the event stream $\mathcal{E}$. Therefore, given a time period $\mathcal{T} = [\tau_{\tsf{start}}, \tau_{\tsf{end}}]$, we construct the graph data as the snapshot $G(\mathcal{T}) = (V(\mathcal{T}), E(\mathcal{T}))$ of the underlying temporal graph. Since there may exist multiple transactions between the same set of users, we consider $G(\mathcal{T})$ to be a \emph{directed multigraph} with each edge in the edge multiset $ E(\mathcal{T}) $ represents an event that happens inside the time interval $\mathcal{T}$, and the node set $ V(\mathcal{T}) $ consists of related users corresponding to the included events. During the training stage, we construct a snapshot $G(\mathcal{T}_{\tsf{train}})$, and obtain a possibly incomplete set of edge labels that are understood as edge properties annotated using expert knowledge. During testing stage, we perform inference over snapshots $G(\mathcal{T}_{\tsf{test}})$ that are based on later time intervals than $\mathcal{T}_{\tsf{train}}$. We assume $\mathcal{T}_{\tsf{train}} \cap  \mathcal{T}_{\tsf{test}} = \emptyset $, hence the problem of interest could be viewed as \emph{inductive edge classification over temporal graphs}. 
    % In the remainder of this section, we introduce our modeling framework as an instantiation of a novelly designed neural message passing protocol that excels in edge-property prediction tasks, with an extension over temporal graphs. 
    
    \subsection{Message passing protocols and directed graphs}
    Let $ G = (V, E) $ be a directed multigraph with node set $ V $ and edge multiset $ E $. For any pairs of nodes $ (u, v) $, denote $ \mu(u, v) $ as the number of edges going from $ u $ to $ v $. Then $ G $ becomes a (simple) graph when $ \max_{u \in V, v \in V} \mu(u, v) \le 1 $. For each $ v \in V $, denote $ N^+(v) = \{u, (v, u) \in E\} $ and $ N^-(v) = \{u, (u, v) \in E\} $ as its out-neighborhood and in-neighborhood respectively, and let $ N(v) = N^+(v) \cup N^-(v) $ be its neighborhood. For the sake of presentation clarity, we will overload the notation $uv$ for both an edge in the undirected graph, or a directed edge from $u$ to $v$ in a directed (multi)graph from time to time, with its exact meaning being clear from the context. We are interested in the general case where both node features $ X = \{x_v\}_{v \in V} $ and edge features $ Z = \{z_{uv}\}_{(u, v) \in E} $ are available, where we assume both kinds of features to be of dimension $d$. In this paper we focus on neural approaches to such directed multigraphs. 
    A good starting point is the neural message passing scheme for undirected graphs \cite{pmlr-v70-gilmer17a}: let $ h^{(l)}_v $ denote the hidden representation of node $ v $ at the $ l $-th layer of the network, and $ h^{(0)}_v = x_v, \forall v \in V $. The message passing graph neural network protocol (abbreviated as GNN hereafter) is described recursively as:
    \begin{align}\label{eqn:gnn_undirected}
        \resizebox{0.91\linewidth}{!}{$
        h^{(l+1)}_v = \tsf{COMBINE}\left( h^{(l)}_v, \tsf{AGG}\left(\tsf{MESSAGE}(h^{(l)}_v, h^{(l)}_u, z_{uv}, u \in N(v))\right)\right)
        $}
    \end{align}
    Different combinations of \tsf{COMBINE}, \tsf{AGG} and \tsf{MESSAGE} mechanisms thus form the \emph{design space} of undirected GNNs \cite{you2020designspace}. 
    To the best of our knowledge, there are three types of generalization strategies to directed graphs:
    \begin{description}
        \item[Symmetrization] The most ad-hoc solution is to  "make it undirected" via padding necessary reverse edges so that $ N(v) = N^+(v) = N^-(v) $, and apply standard graph neural networks that operate on undirected graphs like GCN or GAT. Despite its simplicity and clearness, the symmetrization approach discards directional information in the digraph and may raise subtleties when dealing with multigraphs.
        \item[DiGraph-theoretic motivations] A more recent line of work \cite{ma2019spectral, tong2020directed, NEURIPS2020_cffb6e22, zhang2021magnet} drew insights from directed graph theory, especially the spectral branch \cite{chung2005laplacians}. The proposed models are mostly digraph analogs of GCN, without the consideration for edge features, therefore severely limiting the design space of directed message passing GNNs. 
        \item[Directed Protocol] In the seminal work \cite[Algorithm 1]{battaglia2018relational}, the authors proposed a GNN protocol that operates on directed multigraphs with edge features via aggregating messages from only the in-neighborhood, i.e., replacing $ N(v) $ in \eqref{eqn:gnn_undirected} with $ N^-(v)) $.
        \footnote{The original version also considered incorporation of a \emph{global node} that aggregates information from the whole graph regardless of the connectivity structure. While such design choice may have some gains in moderate size graphs \cite{xiong2021memory}, it does not scale to large graphs. Therefore we will not consider such design choice in this paper}
        While being a natural extension, such kind of GNN protocol losses information from the outgoing direction of each node. We present a pictorial illustration in figure \ref{fig:digraph_illustration}. 
    \end{description}
    To address the aforementioned shortcomings, we propose a novel GNN protocol that operates on directed digraphs termed \emph{bi-directional message passing neural network (BiMPNN)}. The protocol extends the standard undirected protocol \eqref{eqn:gnn_undirected} via enabling each node to aggregate information from both its in-neighborhood and out-neighborhood:
    \begin{align}\label{eqn:bimpnn}
        \resizebox{0.91\linewidth}{!}{$
        \begin{aligned}
            h^{(l+1)}_v &= \tsf{MERGE}\left(\phi^{(l+1)}_v, \psi^{(l+1)}_v\right) \\
            \phi^{(l+1)}_v &= \tsf{COMBINE}_{\tsf{in}}\left( h^{(l)}_v,\right.\\ 
            &\quad \left.\tsf{AGG}_{\tsf{in}}\left(\tsf{MESSAGE}_{\tsf{in}}(h^{(l)}_v,
            h^{(l)}_u, \blue{z_{uv}}, u \in N^-(v))\right)\right)\\
            \psi^{(l+1)}_v &= \tsf{COMBINE}_{\tsf{out}}\left( h^{(l)}_v,\right.\\
            &\quad \left.\tsf{AGG}_{\tsf{out}}\left(\tsf{MESSAGE}_{\tsf{out}}(h^{(l)}_v, h^{(l)}_r, \blue{z_{vr}}, r \in N^+(v))\right)\right)
        \end{aligned}
        $}
    \end{align}
    Despite being more complicated than the undirected protocol, the proposed BiMPNN protocol remains conceptually clear: for each layer, we aggregate separately from each node's in-neighborhood and out-neighborhood using distinct aggregation mechanisms, and merge the two obtained intermediate representations into the next layer's input. Consequently, a $ k $-layer GNN derived from the BiMPNN protocol utilizes both its \emph{root}-$k$ incoming subtree and outgoing subtree, thereby providing a richer set of \emph{relational inductive bias} than the one proposed in \cite{battaglia2018relational}. 
    \subsection{Edge-level task and edge-to-node duality}
    The BiMPNN protocol \eqref{eqn:bimpnn} provides a principled way of obtaining node representations in directed multigraphs which serves as the building block of \emph{node-level tasks}. Yet another important type of graph-related tasks (in a \emph{local} sense \cite{Scarselli2009}) is \emph{edge-level} tasks, which exhibits a dichotomy between \emph{edge-existence} prediction, i.e., link prediction, and \emph{edge-property} prediction, i.e., edge classification. 
    % A notable fact about the latter is that edge features could be incorporated into the modeling procedure, i.e., via formulating the edge-property predictor to be a learnable combination of edge representation, and node representations corresponding to both ends of the edge. 
    For the later task type, the very existence of an edge itself suggests basing the predictions on a properly defined \emph{edge representation}, which should go beyond naively concatenating node representations of its ends \cite{jo2021edge}. Although the protocol \eqref{eqn:bimpnn} implicitly encodes edge features into node representations, it ignores the cascading dynamics of edges (i.e., information implied by cash flow in FRM applications). 
    To build powerful edge representations that efficiently adapt to the underlying graph structure. Mechanisms based on the edge-to-node dual graphs, or \emph{line graph}s, have been proposed \cite{chen2017supervised, jiang2020censnet, cai2020line, jo2021edge} under the undirected GNN protocol \eqref{eqn:gnn_undirected}. To begin our discussion on possible extensions to directed multigraphs, we first review the definition of line graphs as follows:
    \begin{definition}[Line graph and Line digraph \cite{godsil2001algebraic, bang2008digraphs}]
        For both undirected graph and directed (multi)graphs where we overload notation without misunderstandings, $ G = (V, E) $, the node set of its line graph $ L(G) = (L(V), L(E)) $ is defined as its edge (multi)set $ L(V) = E $.\\
        \textbf{Undirected graph} the edge set of its line graph is defined as
        \begin{align}\label{eqn:line}
            \resizebox{0.91\linewidth}{!}{$
            L(E) = \left\lbrace (uv, rs): (u, v) \in E, (r, s) \in E, \{u, v\} \cap \{r, s\} \ne \emptyset \right\rbrace
            $}
        \end{align}
        \textbf{Directed (multi)graph} the edge set of its line graph is defined as
        \begin{align}\label{eqn:diline}
            L(E) = \left\lbrace (uv, rs): (u, v) \in E, (r, s) \in E, v = r\right\rbrace
        \end{align}
    \end{definition}
    For undirected graphs, their line graphs provides a natural way to update edge representations under standard message passing protocols like \eqref{eqn:gnn_undirected}. However, trivially extending \eqref{eqn:bimpnn} using the definition of line digraphs may incur significant information loss: We take the graph in figure \ref{fig:digraph_illustration} as an example, its line graph has an \emph{empty} edge set, which makes the message passing framework useless over the derived line graph. While the graph-theoretic definition enjoys some nice properties \cite{bang2008digraphs}, the adjacency criterion might be overly stringent for deriving useful GNN architectures. Intuitively, we may expect different transactions triggered by the same account as correlated rather than independent, which makes connectivity of edges like $ (n_5, n_9) $ and $ (n_5, n_0) $ as desirable. Therefore, we propose the following \emph{augmentation strategy} to obtain an \emph{augmented edge adjacency graph} $ \overline{L(G)} = (\overline{L(V)}, \overline{L(E)}, T(E)) $: The node set is still defined as $ \overline{L(V)} = E $, and we augment the edge set using the undirected adjacency criterion \eqref{eqn:line}. To retain directional information, we encode the adjacency pattern of two edges (with four possible patterns: \emph{head-to-head}, \emph{head-to-tail}, \emph{tail-to-head}, \emph{tail-to-tail}) into a categorical vector, which we denote as $ T(E) = \{ \tsf{type}(uv, rs): (uv, rs) \in \overline{L(E)} \} $. By construction, for each edge in $ \overline{L(E)} $, its reverse is also in $ \overline{L(E)} $ with possibly different edge types. We provide a pictorial illustration in the left part of figure \ref{fig:linegraphs}. \par 
    To derive an edge representation update rule, we follow the spirit of the BiMPNN node update rule \eqref{eqn:bimpnn}: let $ N^+_L(uv), N^-_L(uv) $ be the out and in neighborhoods in the ordinary line graph $ L(G) $ of $ G $, and $ \overline{N^+_L(uv)}, \overline{N^-_L(uv)} $ to be those in $ \overline{L(G)} $, respectively. For each edge $ (uv, rs) \in \overline{L(E)} $, we use $ \tsf{C}(uv, rs) \in V $ as the common incident node of the edges $uv$ and $rs$. The following updating rule enhances BiMPNN protocol with duality information, which we term BiMPNN-DE:
    \begin{align}\label{eqn:bimpnnde}
        \resizebox{0.91\linewidth}{!}{$
        \begin{aligned}
            h^{(l+1)}_v &= \tsf{MERGE}^{\tsf{node}}\left(\phi^{(l+1)}_v, \psi^{(l+1)}_v\right) \\
            g^{(l+1)}_{uv} &= \tsf{MERGE}^{\tsf{edge}}\left(\theta^{(l+1)}_{uv}, \gamma^{(l+1)}_{uv}\right) \\
            \phi^{(l+1)}_v &= \tsf{COMBINE}^{\tsf{node}}_{\tsf{in}}\left( h^{(l)}_v, \right.\\
            &\quad \left.\tsf{AGG}^{\tsf{node}}_{\tsf{in}}\left(\tsf{MESSAGE}^{\tsf{node}}_{\tsf{in} }(h^{(l)}_v, h^{(l)}_u, \blue{g^{(l)}_{uv}}, u \in N^-(v))\right)\right)\\
            \psi^{(l+1)}_v &= \tsf{COMBINE}^{\tsf{node}}_{\tsf{out}}\left( h^{(l)}_v,\right.\\  
            &\quad \left.\tsf{AGG}^{\tsf{node}}_{\tsf{out}}\left(\tsf{MESSAGE}^{\tsf{node}}_{\tsf{out}}(h^{(l)}_v, h^{(l)}_r, \blue{g^{(l)}_{vr}}, r \in N^+(v))\right)\right) \\
            \theta^{(l+1)}_{uv} &= \tsf{COMBINE}^{\tsf{edge}}_{\tsf{in}}\left( g^{(l)}_{uv},\right.\\  
            &\quad \left.\tsf{AGG}^{\tsf{edge}}_{\tsf{in}}\left(\tsf{MESSAGE}^{\tsf{edge}}_{\tsf{in}}(g^{(l)}_{uv}, g^{(l)}_{pq}, \blue{\hat{h}^{(l)}_{pq, uv}}, pq \in \overline{N_L^-(uv)})\right)\right)\\
            \gamma^{(l+1)}_{uv} &= \tsf{COMBINE}^{\tsf{edge}}_{\tsf{out}}\left( g^{(l)}_{uv},\right.\\  
            &\quad \left.\tsf{AGG}^{\tsf{edge}}_{\tsf{out}}\left(\tsf{MESSAGE}^{\tsf{edge}}_{\tsf{out}}(g^{(l)}_{uv}, g^{(l)}_{rs}, \blue{\hat{h}^{(l)}_{uv, rs}}, rs \in \overline{N_L^+(uv)})\right)\right) \\
            \hat{h}^{(l)}_{uv, rs} &= \tsf{COMBINE}^{\tsf{type}}\left(h^{(l)}_{\tsf{C}(uv, rs)}, T_{\tsf{type}(uv, rs)}\right)
        \end{aligned}
        $}
    \end{align}
    Where we use $ g^{(l)}_{uv} $ to denote the hidden representation of edge $uv$ at the $l$-th layer of GNNs derived from the BiMPNN-DE protocol. The protocol devices an additional edge representation update component that mirrors the BiMPNN protocol over the augmented edge adjacency graph $ \overline{L(G)} $ (see the last four equations in the display \eqref{eqn:bimpnnde}). To obtain an edge representation counterpart $\tilde{h}^{(l)}_{uv, rs}$ during the aggregation process over $ \overline{L(G)} $, we use an additional $\tsf{COMBINE}^{\tsf{type}}$ mechanism that combines features of the common incident node and the information of adjacent types, with is encoded into a learnable edge type embedding matrix $ T \in \mathbb{R}^{4 \times d} $. The BiMPNN-DE protocol \eqref{eqn:bimpnnde} offers a much larger design space than that of BiMPNN protocol. In its full generality, we may specify up to $15$ different mechanisms corresponding to different \tsf{MERGE}, \tsf{COMBINE}, \tsf{AGG} and \tsf{MESSAGE} operations. From a practical point of view, we may design the aforementioned operations using parameterized functions that share the same underlying structure. 
    \subsection{The GRANDE architecture}
    In this section, we devise the previously developed BiMPNN-DE protocol \eqref{eqn:bimpnnde} to derive a concrete GNN architecture that simultaneously outputs node and edge representations, along with an improvement strategy that targets edge-property prediction tasks. We base our design upon the acclaimed Transformer architecture \cite{vaswani2017attention}, which has seen abundant adaptations to GNNs recently \cite{xu2020inductive, dwivedi2020generalization, ying2021transformers}. We define the multiplicative attention mechanism that incorporates edge information as follows:
    \begin{align}\label{eqn:attention}
        \resizebox{0.91\linewidth}{!}{$
        \begin{aligned}
            \tsf{ATTN}&(h_v, \{h_u, g_{uv}\}_{u \in N(v)})= \sum_{u \in N(v) \cup \{v\}} \alpha_{uv} W_N h_u + \beta_{uv} W_E g_{uv} \\
            \alpha_{uv} &= \dfrac{\exp\left(\langle W_Q h_v, W_K h_u\rangle / \sqrt{d} \right)}{\sum_{u \in N(v) \cup \{v\}}\exp\left(\langle W_Q h_v, W_K h_u\rangle/ \sqrt{d}\right)} \\
            \beta_{uv} &= \dfrac{\exp\left(\langle W_Q h_v, W_E g_{uv}\rangle/ \sqrt{d}\right)}{\sum_{u \in N(v) \cup \{v\}}\exp\left(\langle W_Q h_v, W_E g_{uv} \rangle/ \sqrt{d}\right)}
        \end{aligned}
        $}
    \end{align}
    We include commonly used operations in a transformer block, namely LayerNorm (\tsf{LN}), skip connection and a learnable two layer MLP (\tsf{FF}) as nonlinearity \cite{vaswani2017attention}, and wraps them into a transformer block:
    \begin{align}\label{eqn:transformer}
        \resizebox{0.91\linewidth}{!}{$
        \begin{aligned}
            \tsf{TRANSFORMER}&(h_v, \{h_u, g_{uv}\}_{u \in N(v)}) = \tsf{LN}(\tilde{h}_v + \tsf{FF}(\tilde{h}_v))\\
            \tilde{h}_v &= \tsf{LN}(h_v + \tsf{ATTN}(h_v, \{h_u, g_{uv}\}_{u \in N(v)}))
        \end{aligned}
        $}
    \end{align}
    After defining the basic mechanisms, we write the node and edge update rules as follows:
    \begin{align}\label{eqn:grande_base}
        \resizebox{0.91\linewidth}{!}{$
        \begin{aligned}
            h^{(l+1)}_v &= \tsf{CONCAT} \left(\phi^{(l+1)}_v,  \psi^{(l+1)}_v\right) \\
            g^{(l+1)}_{uv} &= \tsf{CONCAT} \left(\theta^{(l+1)}_{uv},  \gamma^{(l+1)}_{uv}\right) \\
            \phi^{(l+1)}_v &= \tsf{TRANSFORMER}^{\tsf{node}}_{\tsf{in}}\left(\Phi_N h_v^{(l)}, \{\Phi_N h_u^{(l)}, \Phi_E g^{(l)}_{uv}\}_{u \in N^-(v)}\right) \\
            \psi^{(l+1)}_v &= \tsf{TRANSFORMER}^{\tsf{node}}_{\tsf{out}}\left(\Psi_N h_v^{(l)}, \{\Psi_N h_r^{(l)}, \Psi_E g^{(l)}_{vr}\}_{r \in N^z=(v)}\right) \\
            \theta^{(l+1)}_{uv} &= \tsf{TRANSFORMER}^{\tsf{edge}}_{\tsf{in}}\left(\Theta_N g_{uv}^{(l)}, \{\Theta_N g_{uv}^{(l)}, \Theta_E \hat{h}^{(l)}_{pq, uv}\}_{pq \in \overline{N_L^-(uv)}}\right) \\
            \gamma^{(l+1)}_{uv} &= \tsf{TRANSFORMER}^{\tsf{edge}}_{\tsf{out}}\left(\Gamma_N g_{uv}^{(l)}, \{\Gamma_N g_{uv}^{(l)}, \Gamma_E \hat{h}^{(l)}_{uv, rs}\}_{rs \in \overline{N_L^+(uv)}}\right)
        \end{aligned}
        $}
    \end{align}
    The updating equations \eqref{eqn:grande_base} involve many learnable parameters, to which we apply the following naming convention: We use upper case greek letters $\Phi, \Psi, \Theta, \Gamma$ to denote projection matrices that takes value in $ \mathbb{R}^{2d \times d} $, and we use the subscript $N$ for \emph{node-related projection} and $E$ for \emph{edge-related projection}. The $\tsf{COMBINE}^{\tsf{type}}$ operation is set to be element-wise addition. \par\noindent
    \textbf{Improvements over edge-property prediction tasks} The architecture in \eqref{eqn:grande_base} is most helpful when the underlying task is to predict properties of some existent edges, which serves as the underlying task for many financial applications like fraud detection and AML. Toward this goal, an ad-hoc solution is to base the prediction with respect to edge $ (u, v) $ upon the concatenation of $h_u$, $h_v$ and $g_{uv}$. According to recent practices over pairwise learning on graphs \cite{wang2019nan, wang2021pairwise}, incorporation of interactions between $ N(u) $ and $ N(v) $, or \emph{cross node interactions} significantly improves prediction performance. However, the cross-node attention module proposed in \cite{wang2019nan} requires full attention between $ h_{N(u)} $ and $ h_{N(v)} $, yielding a potentially large computation overhead. Here we provide a more efficient alternative to model the cross-node interactions called the \emph{cross-query attention} module. Given a node pair $ (u, v) $, compute 
    % \vspace{-1mm}
    \begin{align}\label{eqn:cross_query_attn}
        % \resizebox{0.9\linewidth}{!}{$
        \begin{aligned}
            \delta_{uv} = \tsf{CONCAT}&\left(\tsf{ATTN}_{\tsf{left}}^{\tsf{in}}(h_u, \{h_r, g_{rv}\}_{r \in N^-(v)}), \right. \\
            &\left.\tsf{ATTN}_{\tsf{left}}^{\tsf{out}}(h_u, \{h_s, g_{vs}\}_{s \in N^+(v)})\right) \\
            \delta_{vu} = \tsf{CONCAT}&\left(\tsf{ATTN}_{\tsf{right}}^{\tsf{in}}(h_v, \{h_r, g_{ru}\}_{r \in N^-(u)}), \right. \\
            &\left. \tsf{ATTN}_{\tsf{right}}^{\tsf{out}}(h_v, \{h_s, g_{us}\}_{s \in N^+(u)})\right)
        \end{aligned}
        % $}
    \end{align}
    % \vspace{-1mm}
    The above procedure \eqref{eqn:cross_query_attn} performs four queries that attends $u$ and $v$ to its opponent's in and out neighbors, respectively. Providing the existence of an edge $(u, v)$, the mechanism \eqref{eqn:cross_query_attn} could be understood as attending nodes to the a specific subsets of their \emph{second-order} neighborhoods which have close relationship to the edges of interest. Finally we summarize the previous developments into a framework termed 
    multi\textbf{G}raph t\textbf{RAN}sformer with \textbf{D}uality \textbf{E}nhancement (GRANDE)
    , that utilizes the following edge representation for edge-property prediction tasks:
    % \vspace{-1mm}
    \begin{align}
        g^{\tsf{GRANDE}}_{uv} = \tsf{CONCAT}(g_{uv}, h_v, h_u, \delta_{uv}, \delta_{vu})
    \end{align}
    
    \subsection{Extension to temporal graphs}
    According to the formulation in section \ref{sec:problem_formulation}, extending the \grande framework to $G(\mathcal{T})$ requires utilization of the edge-wise timestamp information $ \{t_{uv}, (u, v) \in E \} $ to produce \emph{time-aware node and edge representations}. Inspired by recent developments of time-aware representation learning approaches \cite{xu2019self, xu2020inductive}, we propose an \emph{extended generic time encoding} mechanism that enhances the node and edge update rule of BiMPNN-DE with temporal information. Finally, to better exploit the structure of the temporal graph, we suggest a \emph{pruning strategy} that shrinks the number of edges of $ \overline{L(G_{\mathcal{T}}))} $ by a factor up to two. \par\noindent
    \textbf{Generic time encoding} The functional time encoding (FTE) provides a principled way that modifies the self-attention operation \eqref{eqn:attention} with minimal architecture change:
    \begin{align}\label{eqn:temporal_attention}
        % \resizebox{0.91\linewidth}{!}{$
        \begin{aligned}
            \tsf{TATTN}&(h_v, \{h_u, g_{uv}\}_{u \in N(v)}) \\
            &= \sum_{u \in N(v) \cup \{v\}} \alpha_{uv} W_N \check{h}_{uv} + \beta_{uv} W_E \check{g}_{uv} \\
            \check{h}_{uv} &= \tsf{CONCAT}(h_u, \tsf{TE}(|\Delta t_{uv}|)) \\
            \check{g}_{uv} &= \tsf{CONCAT}(g_{uv}, \tsf{TE}(|\Delta t_{uv}|))
        \end{aligned}
        % $}
    \end{align}
    Where $ \tsf{TE}(|\Delta t_{uv}|) $ is temporal embedding obtained via certain FTE mechanisms. $\alpha_{uv}$s and $\beta_{uv}$s are calculated analogously to the rules in \eqref{eqn:attention}. 
    % \begin{align}
    %     \begin{aligned}
    %         \alpha_{uv} &= \dfrac{\exp\left(\langle W_Q h_v, W_K \check{h}_{uv}\rangle / \sqrt{d} \right)}{\sum_{u \in N(v) \cup \{v\}}\exp\left(\langle W_Q h_v, W_K \check{h}_{uv}\rangle/ \sqrt{d}\right)} \\
    %         \beta_{uv} &= \dfrac{\exp\left(\langle W_Q h_v, W_E \check{g}_{uv}\rangle/ \sqrt{d}\right)}{\sum_{u \in N(v) \cup \{v\}}\exp\left(\langle W_Q h_v, W_E \check{g}_{uv} \rangle/ \sqrt{d}\right)} \\
    %     \end{aligned}
    % \end{align}
    In this paper we will follow the Bochner-type FTE \cite{xu2019self, xu2020inductive}:
    \begin{align}
        \resizebox{0.85\linewidth}{!}{$
        \tsf{TE}(s) = \sqrt{\frac{1}{d}} \left[ \cos(\rho_1 s), \sin(\rho_1 s), \ldots, \cos(\rho_d s), \sin(\rho_d s)\right]
        $}
    \end{align}
    with learnable parameters $(\rho_1, \ldots, \rho_d)$. According to the setup in \cite{xu2020inductive}, timestamps are associated with nodes rather than edges, hence the authors used the time difference $ \Delta t_{uv} = t_u - t_v $. The situation becomes more complicated when timestamps are associated with edges, for which we propose the following modification:
    \begin{align}\label{eqn:edge_time_difference}
        \Delta t_{uv} = t_{uv} - \min_{u \in N(v)}t_{uv}
    \end{align}
    We use FTE based on \eqref{eqn:edge_time_difference} and replace the \tsf{ATTN} component with \tsf{TATTN} component of the node time difference calculation in the node update part of \eqref{eqn:grande_base}. For the edge update part, the ordinary node time difference method applies naturally since edge timestamps act as node timestamps in the dual formulation. \par\noindent
    \textbf{Temporal information and pruning strategies} the proposed line graph augmentation strategy produces an undirected $ \overline{L(G)} $. While this undirected network might provide valuable information in general, in financial risk management scenarios where edges represent \emph{timestamped transactions}, directed adjacencies between transactions are typically of interest: Consider a transaction $ e_{t_0} $ from Bob to Alice, it is reasonable to assume that \emph{downstream} transactions of $ e_{t_0} $ (i.e., transactions $ e_t $ with $ t > t_0 $) will be affected by $ e_{t_0} $, but its upstream transactions shall not be affected. Inspired from this intuition, we introduce a \emph{causal pruning strategy} that applies when temporal information is available in the underlying graph, i.e, for an edge $ e $ that goes from $ u $ to $ v $, we have its time of occurrence $ t_e $. The causal pruning strategy deletes edges in $ \overline{L(G)} $ with the occurrence time of the head node being earlier than that of the tail node. When the causal pruning strategy is applicable, we may prune up to $50\%$ of the edges in $ \overline{L(G)} $. An illustration is provided in the right part of figure \ref{fig:linegraphs}. Note that the proposed strategy is closely related to the construction of causal temporal subgraphs in temporal graph modeling literature \cite{xu2020inductive, ddgcl}.
    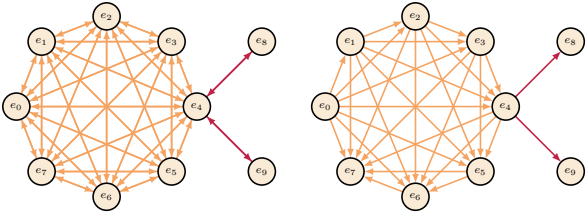
\begin{figure}
        \begin{minipage}{.45\linewidth}
            \centering
            \scalebox{.6}{\begin{tikzpicture}
    \Vertex[label = $ e_0 $, color=antiquewhite, x=-2, y=0]{e0},
    \Vertex[label = $ e_1 $, color=antiquewhite, x=-1.44, y=1.44]{e1},
    \Vertex[label = $ e_2 $, color=antiquewhite, x=0, y=2]{e2},
    \Vertex[label = $ e_3 $, color=antiquewhite, x=1.44, y=1.44]{e3},
    \Vertex[label = $ e_4 $, color=antiquewhite, x=2, y=0]{e4},
    \Vertex[label = $ e_5 $, color=antiquewhite, x=1.44, y=-1.44]{e5},
    \Vertex[label = $ e_6 $, color=antiquewhite, x=0, y=-2]{e6},
    \Vertex[label = $ e_7 $, color=antiquewhite, x=-1.44, y=-1.44]{e7},
    \Vertex[label = $ e_8 $, color=antiquewhite, x=3.44, y=1.44]{e8},
    \Vertex[label = $ e_9 $, color=antiquewhite, x=3.44, y=-1.44]{e9},
    \Edge[Direct, color=sandybrown, lw=1](e0)(e1)
    \Edge[Direct, color=sandybrown, lw=1](e0)(e2)
    \Edge[Direct, color=sandybrown, lw=1](e0)(e3)
    \Edge[Direct, color=sandybrown, lw=1](e0)(e4)
    \Edge[Direct, color=sandybrown, lw=1](e0)(e5)
    \Edge[Direct, color=sandybrown, lw=1](e0)(e6)
    \Edge[Direct, color=sandybrown, lw=1](e0)(e7)
    \Edge[Direct, color=sandybrown, lw=1](e1)(e0)
    \Edge[Direct, color=sandybrown, lw=1](e1)(e2)
    \Edge[Direct, color=sandybrown, lw=1](e1)(e3)
    \Edge[Direct, color=sandybrown, lw=1](e1)(e4)
    \Edge[Direct, color=sandybrown, lw=1](e1)(e5)
    \Edge[Direct, color=sandybrown, lw=1](e1)(e6)
    \Edge[Direct, color=sandybrown, lw=1](e1)(e7)
    \Edge[Direct, color=sandybrown, lw=1](e2)(e0)
    \Edge[Direct, color=sandybrown, lw=1](e2)(e1)
    \Edge[Direct, color=sandybrown, lw=1](e2)(e3)
    \Edge[Direct, color=sandybrown, lw=1](e2)(e4)
    \Edge[Direct, color=sandybrown, lw=1](e2)(e5)
    \Edge[Direct, color=sandybrown, lw=1](e2)(e6)
    \Edge[Direct, color=sandybrown, lw=1](e2)(e7)
    \Edge[Direct, color=sandybrown, lw=1](e3)(e0)
    \Edge[Direct, color=sandybrown, lw=1](e3)(e1)
    \Edge[Direct, color=sandybrown, lw=1](e3)(e2)
    \Edge[Direct, color=sandybrown, lw=1](e3)(e4)
    \Edge[Direct, color=sandybrown, lw=1](e3)(e5)
    \Edge[Direct, color=sandybrown, lw=1](e3)(e6)
    \Edge[Direct, color=sandybrown, lw=1](e3)(e7)
    \Edge[Direct, color=sandybrown, lw=1](e4)(e0)
    \Edge[Direct, color=sandybrown, lw=1](e4)(e1)
    \Edge[Direct, color=sandybrown, lw=1](e4)(e2)
    \Edge[Direct, color=sandybrown, lw=1](e4)(e3)
    \Edge[Direct, color=sandybrown, lw=1](e4)(e5)
    \Edge[Direct, color=sandybrown, lw=1](e4)(e6)
    \Edge[Direct, color=sandybrown, lw=1](e4)(e7)
    \Edge[Direct, color=sandybrown, lw=1](e5)(e0)
    \Edge[Direct, color=sandybrown, lw=1](e5)(e1)
    \Edge[Direct, color=sandybrown, lw=1](e5)(e2)
    \Edge[Direct, color=sandybrown, lw=1](e5)(e3)
    \Edge[Direct, color=sandybrown, lw=1](e5)(e4)
    \Edge[Direct, color=sandybrown, lw=1](e5)(e6)
    \Edge[Direct, color=sandybrown, lw=1](e5)(e7)
    \Edge[Direct, color=sandybrown, lw=1](e6)(e0)
    \Edge[Direct, color=sandybrown, lw=1](e6)(e1)
    \Edge[Direct, color=sandybrown, lw=1](e6)(e2)
    \Edge[Direct, color=sandybrown, lw=1](e6)(e3)
    \Edge[Direct, color=sandybrown, lw=1](e6)(e4)
    \Edge[Direct, color=sandybrown, lw=1](e6)(e5)
    \Edge[Direct, color=sandybrown, lw=1](e6)(e7)
    \Edge[Direct, color=sandybrown, lw=1](e7)(e0)
    \Edge[Direct, color=sandybrown, lw=1](e7)(e1)
    \Edge[Direct, color=sandybrown, lw=1](e7)(e2)
    \Edge[Direct, color=sandybrown, lw=1](e7)(e3)
    \Edge[Direct, color=sandybrown, lw=1](e7)(e4)
    \Edge[Direct, color=sandybrown, lw=1](e7)(e5)
    \Edge[Direct, color=sandybrown, lw=1](e7)(e6)
    \Edge[Direct, color=brightmaroon, lw=1](e4)(e8)
    \Edge[Direct, color=brightmaroon, lw=1](e4)(e9)
    \Edge[Direct, color=brightmaroon, lw=1](e9)(e4)
    \Edge[Direct, color=brightmaroon, lw=1](e8)(e4)
\end{tikzpicture}}
        \end{minipage}
        \begin{minipage}{.45\linewidth}
            \centering
            \scalebox{.6}{\begin{tikzpicture}
    \Vertex[label = $ e_0 $, color=antiquewhite, x=-2, y=0]{e0},
    \Vertex[label = $ e_1 $, color=antiquewhite, x=-1.44, y=1.44]{e1},
    \Vertex[label = $ e_2 $, color=antiquewhite, x=0, y=2]{e2},
    \Vertex[label = $ e_3 $, color=antiquewhite, x=1.44, y=1.44]{e3},
    \Vertex[label = $ e_4 $, color=antiquewhite, x=2, y=0]{e4},
    \Vertex[label = $ e_5 $, color=antiquewhite, x=1.44, y=-1.44]{e5},
    \Vertex[label = $ e_6 $, color=antiquewhite, x=0, y=-2]{e6},
    \Vertex[label = $ e_7 $, color=antiquewhite, x=-1.44, y=-1.44]{e7},
    \Vertex[label = $ e_8 $, color=antiquewhite, x=3.44, y=1.44]{e8},
    \Vertex[label = $ e_9 $, color=antiquewhite, x=3.44, y=-1.44]{e9},
    \Edge[Direct, color=sandybrown, lw=1](e0)(e1)
    \Edge[Direct, color=sandybrown, lw=1](e0)(e2)
    \Edge[Direct, color=sandybrown, lw=1](e0)(e3)
    \Edge[Direct, color=sandybrown, lw=1](e0)(e4)
    \Edge[Direct, color=sandybrown, lw=1](e0)(e5)
    \Edge[Direct, color=sandybrown, lw=1](e0)(e6)
    \Edge[Direct, color=sandybrown, lw=1](e0)(e7)
    \Edge[Direct, color=sandybrown, lw=1](e1)(e2)
    \Edge[Direct, color=sandybrown, lw=1](e1)(e3)
    \Edge[Direct, color=sandybrown, lw=1](e1)(e4)
    \Edge[Direct, color=sandybrown, lw=1](e1)(e5)
    \Edge[Direct, color=sandybrown, lw=1](e1)(e6)
    \Edge[Direct, color=sandybrown, lw=1](e1)(e7)
    \Edge[Direct, color=sandybrown, lw=1](e2)(e3)
    \Edge[Direct, color=sandybrown, lw=1](e2)(e4)
    \Edge[Direct, color=sandybrown, lw=1](e2)(e5)
    \Edge[Direct, color=sandybrown, lw=1](e2)(e6)
    \Edge[Direct, color=sandybrown, lw=1](e2)(e7)
    \Edge[Direct, color=sandybrown, lw=1](e3)(e4)
    \Edge[Direct, color=sandybrown, lw=1](e3)(e5)
    \Edge[Direct, color=sandybrown, lw=1](e3)(e6)
    \Edge[Direct, color=sandybrown, lw=1](e3)(e7)
    \Edge[Direct, color=sandybrown, lw=1](e4)(e5)
    \Edge[Direct, color=sandybrown, lw=1](e4)(e6)
    \Edge[Direct, color=sandybrown, lw=1](e4)(e7)
    \Edge[Direct, color=sandybrown, lw=1](e5)(e6)
    \Edge[Direct, color=sandybrown, lw=1](e5)(e7)
    \Edge[Direct, color=sandybrown, lw=1](e6)(e7)
    \Edge[Direct, color=brightmaroon, lw=1](e4)(e8)
    \Edge[Direct, color=brightmaroon, lw=1](e4)(e9)
\end{tikzpicture}}
        \end{minipage}
        \caption{An illustration of the proposed line graph augmentation strategy: 
        The left figure 
        % stands for the ordinary directed line graph $ L(G) $ of the graph depicted in figure \ref{fig:digraph_illustration}, and the right figure 
        stands for the augmented edge adjacency graph $ \overline{L(G)} $ for the digraph depicted in figure \ref{fig:digraph_illustration}. We use colored edges to represent edge types: {\color{sandybrown} head-to-head} and {\color{brightmaroon} tail-to-tail}, note that the remaining two kinds of edge types do not appear in $ \overline{L(G)} $. The right figure shows the effect of the causal pruning strategy under the additional temporal constraint that $t_0 < t_1 < \cdots < t_9$, with $t_i$ being the occurrence time of edge $t_i$ for $i \in \{0, \ldots, 9\}$}
        \label{fig:linegraphs}
    \end{figure}
    \subsection{Scalability and complexity} 
    Most of the real-world financial networks like transaction networks are \emph{sparse}, i.e., most people only make transactions to a few others given a finite time window. Consequently, the computational complexity of any message passing neural networks could be roughly regarded as $O(Ed^2)$. Extending ordinary MPNN architectures to BiMPNN protocol doubles the computation cost, which could be easily resolved through parallelization in modern deep learning frameworks like tensorflow \cite{abadi2016tensorflow}. The extra computational cost brought by introducing the dual component \eqref{eqn:bimpnnde} requires more care: even when the original graph is sparse, its augmented edge adjacency graph might be dense or even complete. Such cases do happen in realistic scenarios since large hubs frequently exist in transaction networks, which corresponds to a complete subgraph in the dual network. Therefore the worst-case computation cost of $O(E^2d^2)$ is sometimes inevitable in architectures derived from the BiMPNN-DE protocol \eqref{eqn:bimpnnde}. Hence to meet the computational requirement of GNN architectures like \grande, performing GNN training/inference over the whole graph is unrealistic. Instead, we resort to a \emph{local} computation alternative implemented by the AGL system \cite{zhang2020agl} which grabs the $K$-hop rooted subgraph of each target node and performed batched stochastic training and efficient parallel inference given distributed infrastructures like MapReduce \cite{zhang2020agl}. In practical scenarios, it is often reasonable to set an upper bound $M_{\tsf{max}}$ on the edges of any $K$-hop rooted subgraph and device proper sampling methods to meet the requirement. The resulting computational complexity during training is reduced to $O(BM^2_{\tsf{max}}d^2)$, where $B$ denotes the batch size. Since we may control $M_{\tsf{max}}$ so that the whole batch of subgraphs fits the storage requirement of high-performance hardware like GPU, the computational costs of running \grande becomes fully affordable for industry-scale distributed training and inference. 
    \section{Related Works}
    \subsection{Neural models over directed graphs}
    Directional extensions of message passing GNN protocol were mentioned in pioneer works \cite{pmlr-v70-gilmer17a, battaglia2018relational} without providing empirical evaluations. Recent developments toward designing GNNs for digraphs are mostly inspired by different types of graph Laplacians that are defined over digraphs. For example, \cite{NEURIPS2020_cffb6e22} used the definition in  \cite{chung2005laplacians} and \cite{zhang2021magnet} used the Hermitian magnetic Laplacian to decouple the aggregation process of graph connectivity and edge orientations. 
    \subsection{GNNs for edge representation learning}
    The idea of utilizing node-to-edge duality was explored in early works like LGNN \cite{chen2017supervised}, where the authors drew insights from community detection literature, and use the non-backtracking walk operator \cite{krzakala2013spectral} to define the dual graph and perform GCN-like aggregations simultaneously over both graphs. Later developments \cite{jiang2020censnet, cai2020line, jo2021edge} focused on variants of LGNN with the alternative definition of the dual, such as the standard LINE graph \cite{godsil2001algebraic}. 
    \subsection{Transformer architectures over graphs}
    The renowned GAT architecture \cite{velivckovic2017graph} could be regarded as using the \emph{additive} attention mechanism to form the attention layer, as opposed to the \emph{multiplicative} attention mechanism adopted by the Transformer architecture \cite{vaswani2017attention}. Adaptations of the original Transformer to graph context have been assessed recently, \cite{dwivedi2020generalization} replaced the additive attention in GAT with inner product attention and use spectral embedding as a proxy for the positional embedding component in the original transformer architecture. In \cite{ying2021transformers}, the authors proposed to use \emph{full-attention} transformers and use graph-theoretical attributes of nodes and edges to guide the attention procedure. While the results were shown competitive over biological benchmarks, the computational overhead is way too heavy for industrial-level graphical applications.
    \section{Experiments}
    In this section, we report empirical evaluations of \grande over an industrial application as well as assessments over public datasets. We focus on the edge classification task over temporal directed multigraphs. Finally, we present a detailed ablation study to decompose the contributions of different constituents of \grande.
    \subsection{Datasets}
    We use one industrial dataset and two public datasets, with their summary statistics listed in table \ref{tab:dataset_summary} in appendix \ref{sec:app:dataset_summary}.\par\noindent
    \textbf{AML dataset} This dataset is generated from transaction records collected one of the world's leading online payment systems. The business goal is to identify transactions that exhibit risky patterns as being highly suspicious of money laundering. The underlying graph is constructed by treating users as nodes and transactions as directed edges with arbitrary multiplicity. We engineer both node and edge features under a two-stage process: We first obtain raw node features via statistical summaries of corresponding user's behavior on the platform during specific time periods, and raw edge features consist of transaction properties as well as related features of two users involved in the transaction. 
    \footnote{Per organizational regulations, the detailed feature engineering logic is not fully described. We will consider (partially) releasing the AML dataset after passing relevant security checks of the company, as well as the source code.} 
    % After forming the raw features, we apply the decision tree feature transform \cite{he2014practical} using a depth-$6$ complete binary tree that is boosted for $100$ rounds under the gradient boosting scheme \cite{ke2017lightgbm} to both features, via fitting node feature against a list of previously annotated malicious accounts, and edge feature against the edge labels of interest. Consequently, the input node and edge feature for all the assessed models are sparse categorical features with dimension $6400$. 
    The decision tree feature transforms \cite{he2014practical} is then applied to both features so that after the transform, the input node and edge feature for all the assessed models are sparse categorical features with dimension $6400$. For both training and testing, we collect data under a $10$-day period with no overlap between the training period and the testing period. A random subset corresponding to 10\% of the testing data is held out for validation. \par\noindent
    \textbf{Bitcoin datasets} We use two who-trusts-whom networks of people who trade using Bitcoin on two different platforms, Bitcoin OTC and Bitcoin Alpha \cite{kumar2016edge, kumar2018rev2}. Both networks are directed without edge multiplicities, each edge is associated with a timestamp and a trust score ranging from $-10$ to $10$. We consider the task of binary edge classification with edge labels generated as whether the trust score is negative. Using node features represented as the concatenation of one-hot representation of in and out-degree of nodes. For both datasets, we use the chronological split that uses $70\%$ data for training, $10\%$ for validation, and $20\%$ for testing
    \subsection{Baselines}
    We compare the proposed \grande framework with the following types of baselines:
    \begin{description}
        \item[Undirected approaches] We consider two representative GNN architectures GCN \cite{kipf2016semi} and GAT \cite{velivckovic2017graph} that operate on undirected graphs. Since temporal information is available in all three datasets, we also include the TGAT architecture \cite{xu2020inductive}. As all the aforementioned methods produce node-level representations, we use the concatenation of node representations as edge representation according to the adjacency structure. As frameworks that directly output edge representation remain few, we include the EHGNN architecture \cite{jo2021edge} as a strong baseline. To make the undirected architectures compatible with directed (multi)graphs, we add reverse edges with duplicated edge features if there exist no edge multiplicities in the digraph. Otherwise, we keep only one edge between each pair of nodes, with the corresponding edge feature generated via aggregating the original edge features (according to the "multigraph to graph" hierarchy) using the DeepSet method \cite{zaheer2017deepsets}, and add reverse edges thereafter. 
        \item[DiGraph-oriented approaches] We consider two digraph GNN architectures that utilizes different notions of directed graph Laplacians, DGCN \cite{ma2019spectral} and MagNet \cite{zhang2021magnet}. 
    \end{description}
    The aforementioned baselines exclude some of the recently proposed state-of-the-art GNN models like Graphormer \cite{ying2021transformers} for undirected graphs or directed approaches like DiGCN \cite{NEURIPS2020_cffb6e22} due to scalability issues, i.e., they require either full graph attention or solving eigen programs over the full graph Laplacian, which are computationally infeasible for industry-scale graphs. 
    \subsection{Experimental setup}\label{sec: exp_setup}
    Across all the datasets and models, we use a two-layer architecture with hidden dimension $d=128$ without further tuning. For models with generic time encodings, we fix the dimension of time encoding to be $128$. For transformer related architectures, we follow the practice in \cite{vaswani2017attention} and use a two-layer MLP with ReLU activation with hidden dimension $512$. As all the relevant tasks are binary classifications, we adopt the binary cross entropy loss as the training objective, with $\ell_2$ regularization under a coefficient $0.0001$ uniformly across all experiments. The graph data are constructed via the GraphFlat component of the AGL system \cite{zhang2020agl} that transforms the raw graph data into batches of subgraphs with appropriate sampling.
    \footnote{The AGL framework is particularly useful when dealing with industry-scale graphs that are barely possible to process as a whole. However, it may lose some information in the sampling stage of the preprocessing phase. To fully mimic the industrial setup, we preprocess all three datasets using AGL, therefore the results of Bitcoin datasets are not directly comparable to previously published results. }
    We use Adam optimizer with a learning rate of $0.0001$ across all tasks and models. For the bitcoin datasets, we train each model for $10$ epochs using a batch size of $128$ and select the best-performed one according to the roc-auc score on the validation data under periodic evaluations every $100$ steps. For the AML dataset, we train the model for $2$ epochs with a batch size of $256$ as the size of the dataset is sufficiently large. We adopt similar model selection criterion as those of Bitcoin datasets, with periodic evalutions every $500$ steps. \par\noindent
    \textbf{Metrics} Since the primary focus of this paper is applications to the FRM scenario, we choose three representative metrics, namely roc-auc score (AUC), Kolmogorov-Smirnov statistic (KS) and F$1$ score (F$1$). 
    \begin{table*}[]
        \centering
        \caption{Experimental results over two public Bitcoin datasets and the AML dataset, with best performances in \textbf{bold}}
        \begin{tabular}{lccccccccc}
            \toprule
             & \multicolumn{3}{c}{Bitcoin-OTC} & \multicolumn{3}{c}{Bitcoin-Alpha} & \multicolumn{3}{c}{AML}\\
             & AUC & KS & F$1$ & AUC & KS & F$1$ & AUC & KS & F$1$ \\
             \midrule
            GCN \cite{kipf2016semi} & $0.742$ & $0.376$ & $0.432$ & $0.626$ & $0.198$ & $0.282$ & $0.958$ & $0.793$ & $0.704$\\
            GAT 
            \cite{velivckovic2017graph}& $0.736$ & $0.381$ & $0.393$ & $0.626$ & $0.178$ & $0.269$ & $0.962$ & $0.802$ & $0.718$\\
            TGAT \cite{xu2020inductive} & $0.744$ & $0.401$ & $0.422$ & $0.656$ & $0.248$ & $0.294$ & $0.963$ & $0.804$ & $0.720$ \\
            EHGNN \cite{jo2021edge} & $0.719$ & $0.356$ & $0.420$ & $0.626$ & $0.228$ & $0.297$ & $0.961$ & $0.804$ & $0.718$\\
            DGCN \cite{ma2019spectral} & $0.634$ & $0.243$ & $0.354$  & $0.633$ & $0.228$ & $0.282$ & $0.962$ & $0.806$ & $0.720$\\
            MagNet \cite{zhang2021magnet} & $0.753$ & $0.388$ & $0.434$ & $0.645$ & $0.217$ & $0.293$ & $0.954$ & $0.780$ & $0.688$\\
            \midrule
            GRANDE (reduced) & $0.789$ & $0.459$ & $0.460$ & $0.669$ & $0.256$ & $0.294$ &$0.965$ & $0.810$ & $0.726$\\
            GRANDE & \best{0.829} & \best{0.524} & \best{0.532} & \best{0.769} & \best{0.415} & \best{0.398} & \best{0.966} & \best{0.813} & \best{0.734} \\
            \bottomrule
        \end{tabular}
        \label{tab:dataset_results}
    \end{table*}
    \begin{table*}[]
        \centering
        \caption{Performance summary of ablation studies, with best performance in \textbf{bold}}
        \begin{tabular}{lccccccccc}
            \toprule
             & \multicolumn{3}{c}{Bitcoin-OTC} & \multicolumn{3}{c}{Bitcoin-Alpha} & \multicolumn{3}{c}{AML}\\
             & AUC & KS & F$1$ & AUC & KS & F$1$ & AUC & KS & F$1$ \\
             \midrule
            GRANDE & $0.829$ & $0.524$ & \best{0.532} & \best{0.769} & \best{0.415} & \best{0.398} & \best{0.966} & \best{0.813} & \best{0.734} \\
            \midrule
            - reduced & $0.789$ & $0.459$ & $0.460$ & $0.669$ & $0.256$ & $0.294$ &$0.965$ & $0.810$ & $0.726$ \\
            - w/o causal pruning & $0.801$ & $0.464$ & $0.485$ & $0.726$ & $0.325$ & $0.339$ & $0.966$ & $0.813$ & $0.732$ \\
            - w/o time encoding & \best{0.834} & $0.525$ & $0.531$ & $0.691$ & $0.265$ & $0.324$ & $0.966$ & $0.813$ & $0.731$ \\
            - w/o cross-query attention & $0.828$ & \best{0.528} & $0.504$ & $0.766$ & $0.395$ & $0.372$ & $0.966$ & $0.813$ & $0.730$ \\
            - w line graph & $0.762$ & $0.405$ & $0.415$ & $0.655$ & $0.229$ & $0.285$ & $0.965$ & $0.812$ & $0.729$ \\
            \bottomrule
        \end{tabular}
        \label{tab:ablation_study}
    \end{table*}
    \subsection{Performance}
    We present evaluation results in table \ref{tab:dataset_results}. Apart from the proposed \grande architecture, we report a \emph{reduced} version of \grande via discarding all operations on the augmented edge adjacency graph, as well as the cross-query attention module \eqref{eqn:cross_query_attn}. The resulting model could be considered as implementing a time-aware variant of graph transformer under the BiMPNN protocol. We summarize our experimental findings as follows:
    % \vspace{-1mm}
    \begin{itemize}[leftmargin=*]
        \item For the Bitcoin datasets which could be considered as under the \emph{weak feature} regime, the \grande architecture obtains substantial performance improvement: On the Bitcoin-OTC dataset, the relative improvement over the best baselines are $10.1\%$, $30.7\%$ and $22.6\%$ with respect to AUC, KS, and F$1$. On the Bitcoin-Alpha dataset, the relative improvement is more significant with $19.2\%$, $67.3\%$, and $35.4\%$ respectively. We attribute the improvements to both the directional information and the duality information that \grande utilizes. The improvements of the directional information could be inferred from the results of the reduced \grande variant, which exhibits solid improvements over all the baselines. The incorporation of edge-to-node duality and cross-query attention systematically encodes more structural information, therefore yielding further improvements. 
        \item For the AML dataset which could be regarded as under the \emph{strong feature} regime, the performance improvement is significant with respect to KS and F$1$ metrics while being less significant with respect to AUC. Such improvements are still valuable in FRM applications since a higher F$1$ score potentially suggests better patterns of the precision-recall (PR) curve, which we plot in figure \ref{fig:precision_recall}. The PR curve shows the dominant performance of \grande against baselines: under various precision levels, the recall of \grande surpasses the best baseline (TGAT) by as many as $5.29\%$ in absolute value and $13.4\%$ in relative. 
    \end{itemize}
    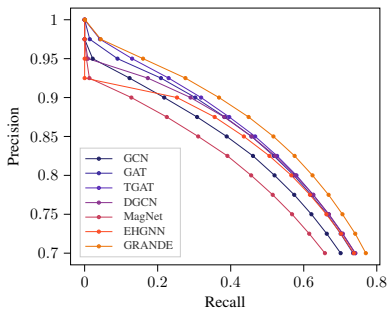
\begin{figure}
        \centering
        \scalebox{.6}{% This file was created with tikzplotlib v0.9.17.
\begin{tikzpicture}

\definecolor{color0}{rgb}{0.117647058823529,0.117647058823529,0.392156862745098}
\definecolor{color1}{rgb}{0.216470588235294,0.15,0.61078431372549}
\definecolor{color2}{rgb}{0.34,0.156666666666667,0.716666666666667}
\definecolor{color3}{rgb}{0.537647058823529,0.189607843137255,0.551960784313726}
\definecolor{color4}{rgb}{0.780392156862745,0.222549019607843,0.342156862745098}
\definecolor{color5}{rgb}{0.985882352941177,0.285294117647059,0.128823529411765}
\definecolor{color6}{rgb}{0.92,0.45,0.03}

\begin{axis}[
legend cell align={left},
legend style={
  fill opacity=0.8,
  draw opacity=1,
  text opacity=1,
  at={(0.03,0.03)},
  anchor=south west,
  draw=white!80!black,
  nodes={scale=0.75, transform shape}
},
tick align=outside,
tick pos=left,
x label style={at={(axis description cs:0.5,-0.1)},anchor=north},
y label style={at={(axis description cs:-0.15,.5)},rotate=90,anchor=south},
x grid style={white!69.0196078431373!black},
xlabel={Recall},
xmin=-0.0385, xmax=0.8085,
xtick style={color=black},
y grid style={white!69.0196078431373!black},
ylabel={Precision},
ymin=0.685, ymax=1.015,
ytick style={color=black},
ytick={0.7, 0.75, 0.8, 0.85, 0.9, 0.95, 1.0}
]
\addplot [semithick, color0, mark=*, mark size=1, mark options={solid}]
table {%
0 1
0 0.975
0.022 0.95
0.123 0.925
0.218 0.9
0.308 0.875
0.389 0.85
0.461 0.825
0.52 0.8
0.574 0.775
0.621 0.75
0.663 0.725
0.701 0.7
};
\addlegendentry{GCN}
\addplot [semithick, color1, mark=*, mark size=1, mark options={solid}]
table {%
0 1
0.014 0.975
0.09 0.95
0.209 0.925
0.303 0.9
0.384 0.875
0.457 0.85
0.518 0.825
0.573 0.8
0.62 0.775
0.664 0.75
0.701 0.725
0.735 0.7
};
\addlegendentry{GAT}
\addplot [semithick, color2, mark=*, mark size=1, mark options={solid}]
table {%
0 1
0.042 0.975
0.13 0.95
0.23 0.925
0.319 0.9
0.396 0.875
0.467 0.85
0.527 0.825
0.58 0.8
0.626 0.775
0.668 0.75
0.706 0.725
0.741 0.7
};
\addlegendentry{TGAT}
\addplot [semithick, color3, mark=*, mark size=1, mark options={solid}]
table {%
0 1
0 0.975
0.009 0.95
0.173 0.925
0.29 0.9
0.382 0.875
0.457 0.85
0.522 0.825
0.577 0.8
0.626 0.775
0.669 0.75
0.707 0.725
0.741 0.7
};
\addlegendentry{DGCN}
\addplot [semithick, color4, mark=*, mark size=1, mark options={solid}]
table {%
0 1
0 0.975
0.005 0.95
0.013 0.925
0.128 0.9
0.225 0.875
0.311 0.85
0.391 0.825
0.456 0.8
0.515 0.775
0.568 0.75
0.615 0.725
0.658 0.7
};
\addlegendentry{MagNet}
\addplot [semithick, color5, mark=*, mark size=1, mark options={solid}]
table {%
0 1
0 0.975
0 0.95
0 0.925
0.253 0.9
0.356 0.875
0.436 0.85
0.506 0.825
0.566 0.8
0.617 0.775
0.662 0.75
0.702 0.725
0.737 0.7
};
\addlegendentry{EHGNN}
\addplot [semithick, color6, mark=*, mark size=1, mark options={solid}]
table {%
0.001 1
0.044 0.975
0.16 0.95
0.276 0.925
0.368 0.9
0.449 0.875
0.517 0.85
0.575 0.825
0.624 0.8
0.668 0.775
0.706 0.75
0.741 0.725
0.77 0.7
};
\addlegendentry{GRANDE}
\end{axis}

\end{tikzpicture}}
        \caption{Precision-recall (PR) curve under the AML dataset. We show recall values corresponding to high precision ranges (from $0.7$ to $1.0$ equally spaced with $0.025$)}
        \label{fig:precision_recall}
    \end{figure}

    \subsection{Ablation study}
    % In this section we conduct further investigations over different components of the \grande architecture. More specifically, we consider the following variants:
    We evaluate the following variants of \grande over all three datasets to investigate contributions of different constituents:
    \begin{description}
        \item[Reduced version] this is the one reported in table \ref{tab:dataset_results}
        \item[Without causal pruning] in this model variant we retain the full edge adjacency graph without pruning. Which is computationally heavier than the \grande architecture
        \item[Without time encoding] in this model variant we discard the temporal component of \grande and use the update rule \eqref{eqn:grande_base}
        \item[Without cross-query attention] in this model variant we discard the cross-query attention module \eqref{eqn:cross_query_attn}, and use $ \tsf{CONCAT}(g_{uv}, h_v, h_u) $ as the output embedding for edge $(u, v)$. 
        \item[With line graph] in this model variant, we use the ordinary directed line graph instead of the proposed augmented edge adjacency graph. i.e., we replace $\overline{N_L^+(uv)}$ and $\overline{N_L^-(uv)}$ in \eqref{eqn:grande_base} with $N_L^+(uv)$ and $N_L^-(uv)$, respectively. 
    \end{description}
    \textbf{Results} we report results in table \ref{tab:ablation_study} using the same training configuration and evaluation metrics as in section \ref{sec: exp_setup}. There are a couple of notable observations: Firstly, the causal pruning procedure saves computation as well as improves performance, providing a solid relational inductive bias in temporal graph modeling. Secondly, the incorporation of time encoding and cross-query attention are in general helpful. Finally, using the ordinary line graph performs on par with the reduced model, showing the insufficiency of additional information provided by line digraphs, thereby verifying the necessity of using the augmented edge adjacency graph.
    \section{Discussion}
    % \tbd
    \textbf{Interpretability} The dominating performance of neural approaches comes at the cost of lacking of \emph{model interpretablity}, which is crucial to application scenarios like AML, where outputs of decision making systems tie strongly with regulatory strictures \cite{kute2021aml}. The adoption of neural approaches enjoys better performance than potentially interpretable methods like linear models as well as losing interpretability. Model explanation methods targeting graph neural models are especially challenging \cite{liu2022interpretability} due to the combinatorial nature of the interpretation problem. Off-the-shelf GNN explaining tools (refer to \cite{liu2022interpretability} and references therein) are not yet applicable to neural models over directed graphs, which is a promising and challenging direction for future explorations. \par\noindent
    \textbf{Multi-task adaptations} The representation quality in both node and edge embeddings gives the \grande architecture the possibility to exploit side-information via multi-task learning paradigms \cite{ruder2017overview}. For example it is quite common in FRM scenarios to obtain both a set of user riskiness labels alongside transaction labels. We have taken a trivial adaptation of \grande into multi-task setups using extra node labels via adding a node-level classification loss to the training objective and has shown solid improvement (we report results in appendix \ref{sec:app:further_exp}). Applying more elegant multi-task learning techniques \cite{ruder2017overview} to further exploit the potential of \grande is an interesting direction for future studies.
    \section{Conclusion}
    In this paper we propose a graph representation learning framework for directed multigraphs that prevail in FRM applications. The proposed framework generalizes the acclaimed message passing graph neural network protocol to incorporate directional information, as well as utilizing the edge-to-node dual relationship to further enhance the relational inductive bias with regard to edge property prediction tasks. A concrete architecture named \grande is derived according to the proposed protocol with the transformer architecture being its aggregation mechanism, as well as a cross-query attention module targeting edge-type tasks. The \grande model is generalizable to temporal dynamic graphs via proper generic time encodings along with a pruning strategy. Experimental results over both public and industrial datasets verify the efficacy of the design of \grande. 
    % \vspace{-1mm}
    \bibliographystyle{acm}
    \bibliography{grande_ref}
    % \clearpage
    \appendix
    \section{Dataset summary}\label{sec:app:dataset_summary}
    We list the summary statistics of used datasets in table \ref{tab:dataset_summary}
    % \begin{table*}
    %     \centering
    %     \caption{Summary statistics of the evaluation datasets}
    %     \begin{tabular}{ccccccc}
    %          \toprule
    %          & \# Nodes & \# Edges & \# Positive edges & \# Negative edges & \# Node features & \# Edge features  \\
    %          \midrule
    %          AML &$10268164$ & $13335278$ & $1338425$ & $11996853$ & $6400$ & $6400$ \\
    %          Bitcoin-otc & $5881$ & $35592$ & $3563$ & $32029$ & $12012$ & $-$ \\
    %          Bitcoin-alpha & $3783$ & $24186$ & $1536$ & $22650$ & $15210$ & $-$ \\
    %          \bottomrule
    %     \end{tabular}
    %     \label{tab:dataset_summary}
    % \end{table*}
    \begin{table}[H]
        \centering
        \caption{Summary statistics of the evaluation datasets}
        \begin{tabular}{lccc}
             \toprule
             & Bitcoin-otc & Bitcoin-alpha & AML \\
             \midrule
             \# Nodes & $5881$ & $3783$ & $10268164$ \\
             \# Edges & $35592$ & $24186$ & $13335278$ \\
             \# Positive edges & $3563$ & $1536$ & $1338425$ \\
             \# Negative edges & $32029$ & $22650$ & $11996853$ \\
             \# Node features & $12012$ & $15210$ & $6400$ \\
             \# Edge features & $-$ & $-$ & $6400$ \\
             \bottomrule
        \end{tabular}
        \label{tab:dataset_summary}
    \end{table}
    \section{Additional experiments}\label{sec:app:further_exp}
    We conduct an additional experiment using the AML dataset augmenting with a set of node labels: the labels are obtained via an expert-maintained malicious user list that takes the form of a binary vector indicating whether the user is malicious or not. The labels are mapped separately to both the buyer (the party that sends money) and seller (the party that receives money) of the transactions, so that for each transaction we may have up to three labels. We follow exactly the same training configuration and hyperparameter settings in section \ref{sec: exp_setup}. Comparisons are made under the recall metric under different precisions, which we report in table \ref{tab:multi_task_comparison}. The results imply that the incorporation of side-information yields consistent improvements, especially in recalls at high precision (with a relative improvement of over $10\%$ in \recall{95}), which is wildly considered to be a key factor in the assessment of models in FRM scenarios. 
    \begin{table}[H]
        \centering
        \caption{Performance comparison between \grande and \grande with side-information (node-level labels) over the AML dataset. metric \recall{p} means recall value at precision $p$, with best performance in bold}
        \begin{tabular}{lcccc}
            \toprule
             & \recall{95} & \recall{90} & \recall{85} & \recall{80} \\
            \midrule
            \grande & $0.160$ & $0.368$ & $0.517$ & $0.624$ \\
            \grande(multi-task) & \best{0.186} & \best{0.382} & \best{0.523} & \best{0.629} \\
            \bottomrule
        \end{tabular}
        \label{tab:multi_task_comparison}
    \end{table}
\end{document}